\documentclass{article}

\PassOptionsToPackage{table}{xcolor}

\usepackage[dblblindworkshop, final]{neurips_2026}
\workshoptitle{The 6th Workshop on Mathematical Reasoning and AI}

\usepackage[utf8]{inputenc}
\usepackage[T1]{fontenc}
\usepackage[table]{xcolor}
\usepackage{hyperref}
\usepackage{url}
\usepackage{booktabs}
\usepackage{amsfonts}
\usepackage{nicefrac}
\usepackage{microtype}
\usepackage{graphicx}
\usepackage{subcaption}

\usepackage{amsmath}
\usepackage{amssymb}
\usepackage{mathtools}
\usepackage{amsthm}

\usepackage{multirow}
\usepackage{arydshln}
\usepackage{listings}
\usepackage[most]{tcolorbox}

\newcommand{\eg}{\textit{e.g.}}

\usepackage[capitalize,noabbrev,nameinlink]{cleveref}
\crefname{equation}{Eq.}{Eqs.}
\creflabelformat{equation}{#2\textup{#1}#3}
\crefname{section}{Sec.}{Secs.}
\crefname{figure}{Fig.}{Figs.}
\crefname{table}{Tab.}{Tabs.}
\crefname{algorithm}{Alg.}{Algs.}

\definecolor{Red}{rgb}{0.768,0.054,0.054}
\definecolor{Blue}{rgb}{0.152,0.294,0.925}
\definecolor{Green}{rgb}{0,0.4,0.7}
\definecolor{mygray}{HTML}{F2F2F2}
\definecolor{myred}{HTML}{BA1200}
\definecolor{citeblue}{HTML}{1668B0}
\hypersetup{
  colorlinks=true,
  citecolor=teal,
  linkcolor=Red,
  urlcolor=Green
}

\newcommand{\xbase}{{\color{myred}x_0}}
\newcommand{\xctx}{{\color{citeblue}x_1}}
\newcommand{\modelname}[1]{\textsc{NudgeRL}}
\newtcolorbox{promptbox}[1]{
  colback=yellow!10!white,
  colframe=yellow!50!black,
  title=#1,
  fontupper=\footnotesize
}

\theoremstyle{plain}
\newtheorem{theorem}{Theorem}[section]
\newtheorem{proposition}[theorem]{Proposition}

\theoremstyle{remark}

\crefalias{proposition}{proposition}
\crefalias{lemma}{lemma}
\crefalias{corollary}{corollary}
\crefalias{definition}{definition}
\crefalias{assumption}{assumption}
\crefalias{remark}{remark}

\crefname{definition}{definition}{definitions}
\Crefname{definition}{Definition}{Definitions}
\crefname{theorem}{theorem}{theorems}
\Crefname{theorem}{Theorem}{Theorems}
\crefname{proposition}{proposition}{propositions}
\Crefname{proposition}{Proposition}{Propositions}
\crefname{lemma}{lemma}{lemmas}
\Crefname{lemma}{Lemma}{Lemmas}
\crefname{corollary}{corollary}{corollaries}
\Crefname{corollary}{Corollary}{Corollaries}
\crefname{assumption}{assumption}{assumptions}
\Crefname{assumption}{Assumption}{Assumptions}
\crefname{remark}{remark}{remarks}
\Crefname{remark}{Remark}{Remarks}

\title{Nudging Beyond the Comfort Zone: \\
Efficient Strategy-Guided Exploration for RLVR}

\author{%
    \textbf{Chanuk Lee}\textnormal{\textsuperscript{1*}}
    \quad
    \textbf{Sangwoo Park}\textnormal{\textsuperscript{1*}}
    \quad
    \textbf{Minki Kang}\textsuperscript{1}
    \quad
    \textbf{Sung Ju Hwang}\textsuperscript{1,2}
    \\[0.5em]
    \rm\textsuperscript{1}KAIST
    \quad\textsuperscript{2}DeepAuto.ai
    \\[0.5em]
    \tt
    \{tallyforce, swgger\}@kaist.ac.kr
}

\begin{document}

\maketitle
{
    \renewcommand\thefootnote{*}
    \footnotetext{Equal contribution}
}

\begin{abstract}
Reinforcement learning with verifiable rewards (RLVR) is a scalable paradigm for improving the mathematical reasoning of large language models, but it is fundamentally limited by exploration: the policy can only improve on trajectories it has already sampled.
Sampling more rollouts alleviates this at prohibitive compute cost, while objective-level modifications offer little control over \emph{what} is explored.
We propose \textsc{NudgeRL}, a framework for structured, diversity-driven exploration in RLVR whose core component, \emph{Strategy Nudging}, conditions each rollout on a lightweight strategy-level context, without requiring the context generator to solve the problem itself.
To learn from such exploration, we decompose the advantage into inter- and intra-context terms and add a policy correction term that transfers discovered behaviors back to the base policy.
Across five mathematical reasoning benchmarks, \textsc{NudgeRL} with 8 rollouts matches the strongest GRPO baseline using 32 rollouts with roughly $5\times$ fewer total tokens and $2.9\times$ less training compute.
Our code is available at \url{https://github.com/tally0818/NudgeRL}.
\end{abstract}

\section{Introduction}
\label{sec:intro}

Reinforcement learning with verifiable rewards (RLVR) has emerged as a powerful paradigm for improving the reasoning capabilities of large language models (LLMs)~\citep{team2025kimi,RLVR_INIT}: by leveraging automatically verifiable signals, most naturally the correctness of a final answer in mathematics, methods such as \textit{Group-Relative Policy Optimization} (GRPO)~\citep{GRPO,guo2025deepseek} enable scalable post-training without dense supervision.

Despite this success, RLVR remains fundamentally limited by its ability to explore the space of reasoning trajectories~\citep{RLVR_MS2, RLVR_MS1}: the policy can only improve on trajectories it has already sampled. Scaling the number of sampled rollouts increases the probability of discovering rare ones~\citep{BRORL}, but quickly becomes computationally prohibitive. Meanwhile, existing alternatives, such as objective-level modification~\citep{DAPO} or utilizing privileged information~\citep{POPE}, either give limited control over \emph{what} is explored or rely on supervision that is expensive to scale (see \Cref{sec:related} for details).

\begin{figure}[t]
    \centering
    \includegraphics[width=0.95\linewidth]{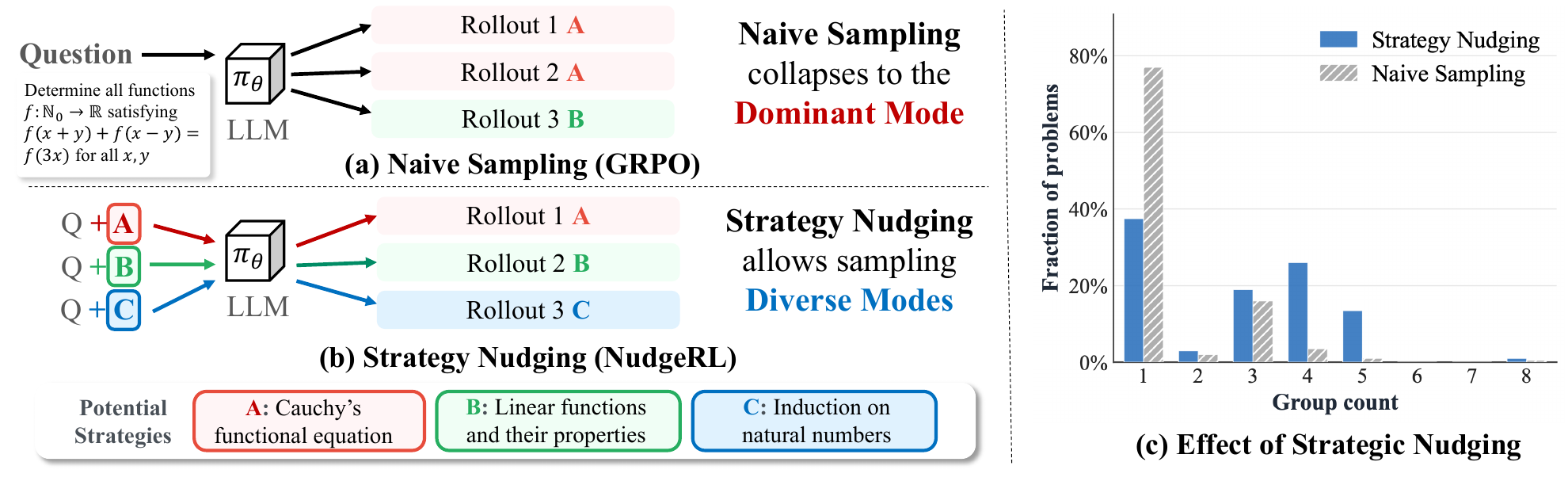}
    % \vspace{-0.1in}
    \caption{\textbf{Improving exploration diversity through Strategy Nudging.} (a) Naive sampling (\eg, GRPO) often collapses to a dominant reasoning mode. (b) \modelname{} appends a lightweight strategy to the input, forcing diverse reasoning modes, which (c) increases the number of distinct reasoning approaches discovered (\Cref{apdx:cf}).}
    \label{fig:concept}
    \vspace{-0.1in}
\end{figure}

We address the exploration bottleneck by explicitly structuring the reasoning space through our proposed framework, \modelname{}. Its core mechanism, \emph{Strategy Nudging}, appends a lightweight strategy-level context (\eg, a single mathematical concept, theorem, or technique such as \textit{``Generating functions''}) during the sampling phase to force the model into exploring distinct reasoning modes it would otherwise ignore under naive sampling. Crucially, unlike prior work using privileged supervision~\citep{POPE}, our context generator only provides high-level guidance and does not need to solve the problem itself, eliminating the labor-intensive process of sourcing oracle solutions.

% \footnote{We use \texttt{gpt-4o-mini} to generate contexts; it yields $\textit{pass@}16=0$ on 33\% of a 200-problem subset of \texttt{DAPO-17k-Processed}.}

Learning from context-conditioned exploration, however, introduces two challenges.
First, because Strategy Nudging assigns varying contexts to different rollouts within the same problem batch, these rollouts form distinct sub-groups. Consequently, their reward variation reflects both trajectory quality and context-specific bias, making standard group-wise advantage estimation unreliable. Second, Strategy Nudging creates a mismatch between how trajectories are sampled and how the policy is used at inference time. We therefore introduce (i) an inter-intra group advantage for credit assignment across context-specific groups, and (ii) an objective augmented with a policy correction term that transfers effective behaviors back to the base policy, detailed in \Cref{sec:interintra} and \Cref{sec:training_objective}, respectively.

\section{Related Work}
\label{sec:related}

RLVR~\citep{RLVR_INIT} is now a mainstream post-training paradigm for improving the reasoning capability of LLMs~\citep{GRPO, DAPO, blakeman2025nvidia}, and its limitations arising from insufficient exploration, such as marginal performance gains~\citep{BRORL} and mode collapse~\citep{RLVR_MS1, RLVR_MS2}, have become apparent. One line of work~\citep{RLVR_ENT, RLVR_ENT3, Passk2, DAPO} promotes exploration at the objective level through entropy bonuses, direct \textit{pass@}$k$ objectives, or decoupled clipping, which affords little control over \textit{what} is explored.
A complementary line~\citep{chen2025nudging, POPE, BREAD, SAGE} leverages privileged information, but is fundamentally \textit{feasibility}-oriented: costly and poorly scalable supervision, such as oracle solutions, is introduced precisely when the policy fails to sample any correct trajectory (extended discussion in \Cref{apdx:related}).

\section{\modelname{}}
\label{sec:method}

\paragraph{Preliminaries.}
For each prompt $x_0$ of an empirical distribution $\mathcal{D}$, a policy $\pi_\theta$ generates $N$ rollouts $y_i\sim\pi_\theta(\cdot\mid x_0)$, each scored by a verifiable reward $r_i = R(x_0,y_i)\in\{0,1\}$.
GRPO~\citep{GRPO} replaces a learned value function with group-wise statistics, setting $\hat A_i=(r_i-\mu)/(\sigma+\delta)$ for the within-group reward mean $\mu$ and standard deviation $\sigma$ ($\delta>0$ for stability), and optimizes a PPO-style clipped objective~\citep{PPO} over $\rho_i = \pi_\theta(y_i|x_0)/\pi_\textrm{old}(y_i|x_0)$.
Because the update only reweights sampled trajectories, RLVR is bottlenecked by the probability mass on \emph{unexplored} correct regions, which scaling $N$ shrinks only slowly for low-probability, long-tail trajectories, making naive scaling highly inefficient (see \Cref{apdx:brorl} for details).

\subsection{Structured Exploration via Strategy-Level Contexts}
\label{sec:strategy-nudging}

To illustrate the need for structured exploration, we consider how many rollouts are required to reliably discover a rare trajectory. For $y$ with $\pi(y\mid x_0)\ll 1$, the expected number needed to observe $y$ at least once is $1/\pi(y\mid x_0)$, which grows prohibitively large since naive scaling repeatedly samples the policy's high-probability modes. Instead, we can explicitly steer the model toward unexplored regions by introducing a strategy-level context $c$. This targeted guidance shifts the sampling distribution such that $\pi(y\mid x_0,c)\gg\pi(y\mid x_0)$ drops the expected discovery cost to $1/\pi(y\mid x_0,c)$.
Contexts therefore need not provide a solution; they are lightweight controls that reduce discovery cost.

\begin{table}[t]
\centering
\caption{
\textbf{Accuracy and training cost} on \texttt{Qwen3-4B-Instruct-2507}, with \textit{pass@}1 estimated using 128 rollouts (best \textbf{bold}, second best \underline{underlined}). Tokens include \modelname{}'s one-time offline context generation, EFLOPs are $10^{18}$ FLOPs, and Gain / B tok.\ is the average \textit{pass@}1 gain over the base model per billion total tokens. \texttt{Olmo3-7B-Instruct-SFT} results are in \Cref{apdx:model}.
}
\label{tab:main_table}
\small
\resizebox{\textwidth}{!}{
\setlength{\tabcolsep}{5.5pt}
\begin{tabular}{l c cccccc ccc}
\toprule
\multirow{2}{*}{\textbf{Method}} & \multirow{2}{*}{\textbf{Rollouts ($N$)}}
& \multicolumn{6}{c}{\textbf{\textit{pass@}1}}
& \multicolumn{3}{c}{\textbf{Training cost / efficiency}} \\

\cmidrule(lr){3-8} \cmidrule(lr){9-11}

& & \textbf{AIME24} & \textbf{AIME25}
& \textbf{AMC23} & \textbf{MATH500} & \textbf{APEX} & \textbf{Average}
& \textbf{Tokens (M)} & \textbf{EFLOPs} & \textbf{Gain / B tok.} \\

\midrule

Base model & --
& 0.374 & 0.352 & 0.653 & 0.592 & 0.036 & 0.402 & -- & -- & -- \\

\cmidrule{1-11}

\multirow{4}{*}{GRPO}
& 8
& 0.444 & 0.367 & 0.749 & \underline{0.668} & 0.040 & 0.454 & \underline{10.8} & \textbf{0.342} & \underline{4.83} \\
& 16
& 0.454 & 0.355 & 0.840 & 0.655 & 0.045 & 0.470 & 22.0 & 0.701 & 3.09 \\
& 32
& 0.451 & \underline{0.370} & \textbf{0.881} & \textbf{0.674} & \textbf{0.058} & \underline{0.487} & 44.7 & 1.421 & 1.90 \\
& 64
& 0.415 & 0.324 & 0.848 & 0.641 & 0.027 & 0.451 & 87.6 & 2.786 & 0.56 \\

\cmidrule{1-11}

\rowcolor{citeblue!10}
\modelname{} & 8
& \textbf{0.482} & \textbf{0.393} & \underline{0.857} & 0.660 & \underline{0.053} & \textbf{0.489} & \textbf{9.0} & \underline{0.492} & \textbf{9.68} \\

\bottomrule
\end{tabular}
}
\vspace{-0.05in}
\end{table}

\paragraph{Strategy Nudging.}
Placing all contexts in a single prompt leaves the choice of strategy to the policy, which may ignore some and repeat dominant patterns, so we assign a single sampled context to each rollout before generation.
Let $\mathcal{C}(x_0)=\{c_1,\ldots,c_M\}$ be a pool of strategy-level contexts for prompt $x_0$, from which we sample $c^{(i)}\sim\mathrm{Uniform}(\mathcal{C}(x_0))$ for each rollout index $i$.
To avoid relying exclusively on the pool and to retain compatibility with the original prompt, we apply context dropout: we sample $b^{(i)}\sim\mathrm{Bernoulli}(1-p_{\mathrm{drop}})$ and set $z^{(i)}=c^{(i)}$ if $b^{(i)}=1$ and $z^{(i)}=\emptyset$ otherwise, then form $x_1^{(i)}=(x_0,z^{(i)})$ and generate $y_i\sim\pi_\theta(\cdot\mid x_1^{(i)})$.
Details on generating $\mathcal{C}$ are in~\Cref{apdx:cf}.

\paragraph{Context-induced rollout diversity.}
We verify this under an identical budget of 8 rollouts per prompt, drawing 4 from each of 2 contexts without dropout while the baseline draws all 8 under the original prompt.
Clustering the reasoning structures with an LLM-as-a-judge (\texttt{gpt-4o-mini}~\citep{gpt4o}), \Cref{fig:concept} shows that Strategy Nudging more often increases the number of distinct reasoning structures, whereas the base policy frequently collapses to similar patterns.

\subsection{Inter-Intra Group Advantage: Balancing Exploration between Strategies}
\label{sec:interintra}

To learn from context-forced exploration, we assign credit through two complementary signals: an \emph{intra}-context signal capturing trajectory quality under the same context, and an \emph{inter}-context signal capturing the relative reliability of the context itself.
Given rollouts $\{y_i\}_{i=1}^N$ with rewards $r_i$, we group them by assigned context, $\mathcal{G}(x_0)=\mathrm{Unique}(\{z^{(i)}\})$, define $I_g=\{i\mid z^{(i)}=g\}$ for each $g\in\mathcal{G}(x_0)$, and set $\bar r_g=\frac{1}{|I_g|}\sum_{i\in I_g} r_i$ and $\bar r=\frac{1}{N}\sum_{i=1}^N r_i$, giving
\begin{equation}
    \hat{A}_i = \frac{A_i - \mu_A}{\sigma_A + \delta},\quad\text{where }
    A_i =
    \begin{cases}
    \underbrace{(r_i - \bar{r}_{z^{(i)}})}_{\text{intra-context}}
    +
    \lambda \underbrace{(\bar{r}_{z^{(i)}} - \bar{r})}_{\text{inter-context}}
    & \text{if } z^{(i)} \neq \emptyset, \\[2pt]
    r_i - \bar{r}
    & \text{if } z^{(i)} = \emptyset,
    \end{cases}
\end{equation}
with $\mu_A,\sigma_A$ the mean and standard deviation of $\{A_i\}$.
Since advantages determine the direction of the update, they should stay consistent with the underlying rewards while still letting context-level preferences affect credit assignment.

\begin{proposition}
\label[proposition]{proposition:3.1}
Let $y$ and $y'$ be trajectories sampled from context groups $z$ and $z'$ with rewards $r$ and $r'$ and advantages $A$ and $A'$. In the binary reward setting, if $\lambda \in (0,2]$, then $r > r' \Rightarrow A > A'$.
\end{proposition}

Thus, for $\lambda\in(0,2]$ a higher reward always receives a higher advantage, and the context only affects the ordering among \emph{equal}-reward trajectories: $\lambda<1$ favors successes from lower-reward contexts, encouraging exploration of less typical ones, $\lambda>1$ emphasizes more reliable ones, and $\lambda=1$ exactly recovers GRPO (proof and an extension to non-binary rewards in \Cref{apdx:reward}).

\subsection{Training Objective}
\label{sec:training_objective}
Since the target policy operates without external contexts at inference time, trajectories discovered under the context-conditioned prompt $\xctx=(\xbase,z)$ must be transferred to $\pi_\theta(\cdot\mid\xbase)$.
We bridge this gap with a policy correction term, adapting advantage-weighted distillation from~\citet{RLTF}:
\begin{equation}
    \mathcal{L}_\textrm{PC}(\theta)
    =
    -\mathbb{E}_{y \sim \pi_\theta(\cdot|\xctx)}
    \big[\hat{A} \log \pi_\theta(y|\xbase)\big].
\end{equation}
Unlike behavior cloning, this emphasizes trajectories with high normalized advantage. In parallel we optimize the clipped RL objective on the \emph{context-conditioned} policy,
\begin{equation}
    \mathcal{L}_\textrm{RL}(\theta)
    =
    -\mathbb{E}_{y \sim \pi_\theta(\cdot|\xctx)}
    \big[
    \min(r_1 \hat{A},\
    \mathrm{clip}(r_1, 1-\epsilon_\textrm{low}, 1+\epsilon_\textrm{high}) \hat{A})
    \big],
\,\text{where}\,
    r_1
    =
    \frac{\pi_\theta(y|\xctx)}
    {\pi_\textrm{old}(y|\xctx)},
\end{equation}
and combine both as $\mathcal{L}_\textrm{\modelname{}} = \mathcal{L}_\textrm{RL} + \lambda_\textrm{PC}\mathcal{L}_\textrm{PC}$: the RL term reinforces successful trajectories within each context, while policy correction projects these improvements onto the base-prompt policy (\Cref{apdx:overview}).

\section{Experiments}
\label{sec:exp}

\paragraph{Setup.}
We apply \modelname{} to \texttt{Qwen3-4B-Instruct-2507}~\citep{Qwen3}, training on \texttt{DAPO-17k-Processed}~\citep{DAPO} with two strategy-level contexts per problem (\eg, \textit{``Pythagorean theorem''}) generated by \texttt{gpt-4o-mini}~\citep{gpt4o} and used without verification.
We compare against the base model and GRPO with rollout budgets $N\in\{8,16,32,64\}$ on \texttt{AIME24}/\texttt{AIME25}, \texttt{AMC23}, the level-5 subset of \texttt{MATH500}, and the \texttt{Apex Shortlist}, reporting \textit{pass@}1 estimated from 128 rollouts~\citep{passk} and grading with \texttt{math-verify}~\citep{mathverify2024}; benchmark references and further details are in \Cref{apdx:baselines,apdx:training,apdx:eval}.

\subsection{Main Results}

As shown in~\Cref{tab:main_table}, \modelname{} attains an average \textit{pass@}1 of 0.489, matching or slightly exceeding the best GRPO configuration (0.487 at $N=32$) with one-fourth of the rollout budget, and outperforming GRPO at $N=64$ (0.451), which highlights the instability of naive rollout scaling~\citep{BRORL}.
The trend holds on \texttt{Olmo3-7B-Instruct-SFT}~\citep{Olmo} (\Cref{apdx:model}) and beyond mathematics on \texttt{LiveCodeBench}~\citep{LiveCodeBench}, where \modelname{} beats GRPO at every budget up to $N=64$ (\Cref{apdx:code}).

\subsection{Analysis on Efficiency}

\paragraph{Token and training efficiency.}
The cost columns of \Cref{tab:main_table} show that \modelname{} reaches its accuracy with $5.0\times$ fewer total tokens and $2.9\times$ fewer training FLOPs than GRPO at $N=32$, even after charging the one-time offline context-generation cost to our method and despite the extra forward and backward pass of policy correction, which alone adds roughly $44\%$ compute at a matched rollout budget.
Notably, the average \textit{pass@}1 gain over the base model per billion tokens consistently \emph{decreases} as the GRPO budget grows, consistent with \Cref{apdx:brorl} and~\citet{isocompute}; per-billion-token gains and the FLOPs protocol are in \Cref{apdx:effie}.

\paragraph{Where the gains come from.}
\modelname{} improves \textit{pass@}1 faster than every GRPO variant during training and dominates GRPO-8 across the full range of inference-time budgets $k$ (\Cref{apdx:dynamics}): Strategy Nudging exposes useful trajectories \emph{earlier} rather than sampling more of them.
On an \texttt{AIME25} case study, \modelname{} amplifies a rare but decisive \emph{shoelace-formula} strategy that GRPO samples only once (\Cref{apdx:case}).
Ablations (\Cref{apdx:ablation,apdx:additional}) show that moderate context dropout ($p_\textrm{drop}=0.5$) is best and that \emph{randomly} sampled contexts beat oracle top-ranked and self-generated ones: a context's role is not to supply the single best hint, but to promote diversity in reasoning.

\section{Conclusion}

We introduced \modelname{}, a framework for structured exploration in RLVR that uses \emph{Strategy Nudging} to induce diverse reasoning trajectories from lightweight strategy-level contexts and learns from them via a policy correction-augmented RL objective, matching the best GRPO configuration with $5\times$ fewer tokens and $2.9\times$ less training compute.

\paragraph{Limitations \& future work.}
\label{sec:limit}
Generating contexts has a cost, though it is an \emph{offline}, one-time process using a lightweight LLM and the contexts are reusable across runs.
More fundamentally, contexts are generated independently of the policy, so a fixed pool may become less informative as training progresses; \emph{model-adaptive context generation} is a promising direction.

\newpage
\bibliography{reference}
\bibliographystyle{plainnat}

\newpage
\appendix
\section{Related Work}
\label{apdx:related}

\subsection{Reinforcement Learning with Verifiable Rewards}

Reinforcement Learning with Verifiable Rewards (RLVR) has emerged as a scalable paradigm for improving the reasoning abilities of large language models~\citep{team2025kimi,GRPO,DAPO,DS,guo2025deepseek}.
By leveraging automatically verifiable signals, such as exact answers in mathematics or test-case correctness in code generation, RLVR enables effective policy optimization without dense human supervision.

A representative approach is \textit{Group-Relative Policy Optimization} (GRPO)~\citep{GRPO,guo2025deepseek}, which replaces value function estimation with group-wise comparisons among sampled rollouts, deriving advantages from relative reward differences within each group.
Building on this formulation, subsequent work has introduced improvements such as decoupled clipping~\citep{DAPO} and alternative normalization strategies~\citep{DRGRPO, zheng2025group} to enhance training stability.

These methods have been successfully applied across a range of reasoning tasks and model scales~\citep{prorl,DS}, establishing RLVR as a standard post-training approach for LLMs.
However, their effectiveness fundamentally depends on exploration: the policy can only improve on trajectories it has already sampled.
As a result, insufficient exploration directly limits learning, making it a key bottleneck in RLVR.
We next examine how prior work addresses this challenge.

\subsection{Exploration in RLVR}
A straightforward approach to improving exploration is to scale the number of sampled rollouts. Prior work has shown that such rollout scaling can significantly improve performance by reducing the probability mass of the un-sampled region~\citep{BRORL}. However, this approach is computationally expensive and often impractical at scale.

More commonly, recent methods attempt to encourage exploration through objective design, such as entropy regularization~\citep{RLVR_ENT, RLVR_ENT3} or decoupled clipping~\citep{DAPO}. While these approaches can steer the update toward exploration, they do not guarantee that useful or rare modes are actually sampled during training. In other words, shaping the distribution does not necessarily ensure coverage of meaningful trajectories, leaving exploration fundamentally limited.

Moreover, such distribution-level exploration is inherently stochastic and unconstrained, which can perturb the policy in semantically undesirable directions. Increasing entropy or aggressively reweighting probabilities may encourage the model to explore low-probability regions, but without any structural guidance, this often leads to incoherent or unproductive trajectories rather than meaningful reasoning strategies. As a result, these approaches lack control over \textit{how} the policy explores, and fail to provide structured, strategy-level exploration that targets diverse and semantically valid modes of reasoning.

\subsection{Usage of Privileged Information}
Another key limitation of widely used group-based advantage methods, such as GRPO~\citep{GRPO}, is that they rely on relative comparisons within a group of rollouts. When all samples in a group are either correct or incorrect, these methods fail to provide informative learning signals.

To address this issue, recent works have introduced privileged information to assist the policy~\citep{BREAD, POPE, RLTF, SAGE, chen2025nudging}, often in the form of oracle prefixes or intermediate solutions. These approaches improve the feasibility of solving hard problems by enabling the model to generate successful trajectories that would otherwise be unreachable.

However, such methods come with several limitations. First, privileged information is often difficult to scale, especially when it relies on oracle solutions or expensive annotations~\citep{BREAD,POPE}. Second, the mechanism by which the model internalizes this information and performs well without it at test time remains unclear~\citep{SAGE}. Third, many approaches assume multi-turn or interactive settings~\citep{RLTF}, which may not align with standard single-turn RLVR setups.

More importantly, existing work primarily focuses on improving the feasibility of generating correct trajectories on difficult problems. In contrast, our work targets a complementary challenge: improving the diversity of exploration, even when successful trajectories are already attainable.

\section{The Exploration Bottleneck: From Exploration to Performance Gain}
\label{apdx:brorl}

To understand why exploration is a fundamental bottleneck in RLVR, we look beyond trajectory-level rewards and examine how the probability mass of generated tokens shifts during training. \citet{BRORL} characterizes the expected one-step performance improvement ($\Delta Q_\text{pos}$) in RLVR as:
\begin{equation}
    \label{eq:brorl}
    \Delta Q_\text{pos}=\frac{\eta}{N}
    \big[(1-S_R)Q_\text{neg}A_2+S_RQ_\text{pos}B_2+S_R(Q_\text{pos}U_{\text{neg},2}-Q_\text{neg}U_{\text{pos},2})\big],
\end{equation}
where $Q_\text{pos}$ and $Q_\text{neg}=1-Q_\text{pos}$ denote the total probability mass of correct and incorrect tokens, $\eta$ is the learning rate, and $N$ is the number of rollouts.
$A_2$ and $B_2$ are the second moments of \emph{sampled} correct and incorrect tokens, while $U_{\text{pos},2}$ and $U_{\text{neg},2}$ are those of \emph{unsampled} correct and incorrect tokens.
$S_R \in [0,1]$ represents the net reward contribution from sampled tokens.

Since $S_R \in [0,1]$, the first two terms in \Cref{eq:brorl} are non-negative and drive learning forward. The third term, however, acts as a potential penalty.
Because incorrect tokens typically dominate the probability mass ($Q_\text{neg} \gg Q_\text{pos}$), a large $U_{\text{pos},2}$, meaning the model has significant probability mass on correct trajectories that it simply \emph{fails to explore}, creates a dominant negative force that hinders performance gain. Therefore, the core bottleneck of RLVR lies in the \emph{unexplored} correct regions.

\paragraph{Limitations of rollout scaling.}
To mitigate this penalty, a naive solution is to increase the rollout size $N$.
\citet{BRORL} shows that for a collection of tokens with probabilities $\{p_i\}$, the expected unsampled second moment after $N$ draws is
\begin{equation}
    \sum_i p_i^2 (1 - p_i)^N,
\end{equation}
which decreases monotonically with $N$.
However, tokens with small $p_i$ decay slowly, so fully covering long-tail correct trajectories requires prohibitively large rollout budgets.

This highlights the limitation of blindly scaling $N$ to reduce the unexplored correct mass ($U_{\text{pos},2}$). Long-tail correct trajectories remain unlikely to be sampled even under large $N$, suggesting the need for a \emph{structured exploration} mechanism that can \emph{efficiently} expose such latent trajectories, which is precisely what Strategy Nudging provides (\Cref{sec:strategy-nudging}).

\section{Overview of the Learning Mechanism}
\label{apdx:overview}

\begin{figure}
    \centering
    \includegraphics[width=0.95\linewidth]{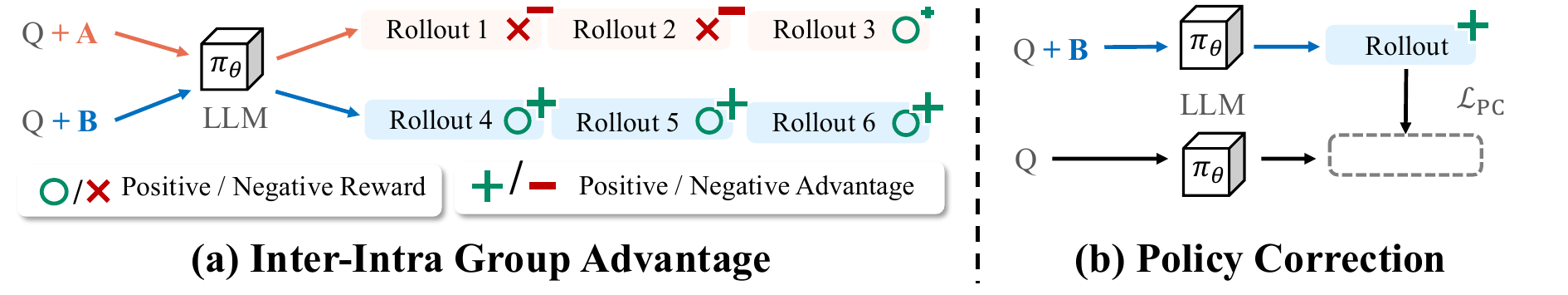}
    \vspace{-0.1in}
    \caption{\textbf{Overview of the NudgeRL learning mechanism.} \textbf{(a) Inter-Intra Group Advantage:} Demonstrates credit assignment that emphasizes reliable contexts (i.e., $\lambda \in (1, 2]$). A successful rollout from a consistently high-reward context (\textcolor{blue}{Strategy B}) receives a larger positive advantage than a rare success from a low-reward context (\textcolor{orange}{Strategy A}). \textbf{(b) Policy Correction:} High-quality trajectories discovered via context-conditioned exploration ($Q+B$) are distilled back into the base policy ($Q$) using $\mathcal{L}_{\text{PC}}$, allowing the model to internalize effective reasoning modes.}
    \label{fig:method}
    \vspace{-0.18in}
\end{figure}

\Cref{fig:method} illustrates the two learning components of \modelname{} introduced in \Cref{sec:interintra} and \Cref{sec:training_objective}: the Inter-Intra Group Advantage, which assigns credit both within and across context groups, and Policy Correction, which distills trajectories discovered under context-conditioned exploration back into the base-prompt policy.

\section{Details on Strategy Nudging}
\label{apdx:cf}
\paragraph{Strategy Generating Prompt.}
We use \texttt{gpt-4o-mini} to generate keyword-level hints for each problem.
For the main experiments, we generate two hints per problem, while in the top-ranked setting, we first generate five candidate hints and select a subset based on oracle evaluation.

The exact prompt used for hint generation is as follows:
\begin{lstlisting}
f"""Given the following math problem, generate {num_hints} different keyword hints that would help solve it. 
Each hint should be a specific mathematical concept, theorem, or technique (e.g., "Ceva's theorem", "Lifting the exponents", "Triangle inequality").

Problem:
{problem}

Please provide exactly {num_hints} hints in the following format (one hint per line, numbered):
1. [Hint 1]
2. [Hint 2]
...
{num_hints}. [Hint {num_hints}]

Make sure each hint is a distinct mathematical concept or theorem."""
\end{lstlisting}

\paragraph{Strategy Nudging prompt.} Given a problem and an optional hint, we construct prompts that encourage the model to follow a specific reasoning strategy. 
The model is instructed to explicitly separate its reasoning process and final answer using predefined delimiters.
\begin{lstlisting}
    reasoning_start = "<start_working_out>"
    reasoning_end = "<end_working_out>"
    solution_start = "<SOLUTION>"
    solution_end = "</SOLUTION>"
    
    system_prompt = f"""You are given a problem.
    Think about the problem and provide your working out.
    Place it between {reasoning_start} and {reasoning_end}.
    Then, provide your solution between {solution_start}{solution_end}"""
    
    def build_messages(problem: str, system_prompt: str, hint: str | None = None) -> list[dict[str, str]]:
        context_block = ""
        if hint:
            context_block = (
                "Context (exploration condition):\n"
                f"- Use this hint/approach: {hint}\n\n"
                "Important:\n"
                "- Follow this approach as your primary strategy.\n\n"
            )
        user_content = (
            "Problem:\n"
            f"{problem}\n\n"
            f"{context_block}"
            "Solve this step by step and provide your final numerical answer at the end."
        )
        return [
            {"role": "system", "content": system_prompt},
            {"role": "user", "content": user_content},
        ]
\end{lstlisting}
\paragraph{Effect of Strategy Nudging.}
To evaluate the effect of Strategy Nudging, we sample 8 rollouts from \texttt{Qwen3-4B-Instruct-2507}~\citep{Qwen3} on 200 problems from \texttt{DAPO-17k-Processed}~\citep{DAPO}, both with and without Strategy Nudging, and analyze the resulting rollout diversity via LLM-as-a-judge.

\paragraph{LLM-as-a-judge prompt.}
To analyze the diversity of generated rollouts, we employ an LLM-as-a-judge using \texttt{gpt-4o-mini} to cluster solutions based on their underlying reasoning strategies and count the number of distinct solution modes.
Given a problem and a set of rollouts, the model is instructed to identify the number of \emph{conceptually distinct} solution approaches, while ignoring superficial differences such as phrasing or minor computational variations.
\begin{lstlisting}
    prompt = (f"Problem:\n{problem_text}\n\n"
              f"Here are {len(rollouts)} student solutions to this problem:\n"
              f"{formatted_rollouts}\n"
              f"Analyze these solutions and determine how many *conceptually distinct* solution strategies are used across them.\n"
              f"Ignore minor calculation differences or phrasing variations. Focus on the core mathematical approach.\n"
              f"Provide the answer in the following format: 'Distinct Strategies: X' where X is the integer count.\n"
              f"Then briefly list the distinct strategies identified.")
    messages = [
        {"role": "system", "content": "You are an expert math teacher evaluating the diversity of student solution methods."},
        {"role": "user", "content": prompt}
    ]
\end{lstlisting}
\section{Further Analysis of $\lambda$}
\label{apdx:reward}

In this section, we provide a detailed interpretation of the proposed advantage estimator, prove \cref{proposition:3.1}, and extend the reward-ordering analysis beyond binary rewards.

For notational convenience, we define the effective group baseline as
\begin{equation}
    b_z
    =
    \begin{cases}
        \bar{r}_z, & z \neq \emptyset,\\
        \bar{r}, & z = \emptyset.
    \end{cases}
    \label{eq:effective_group_baseline}
\end{equation}
The unnormalized advantage of a rollout $y_i$ with reward $r_i$ and assigned context $z^{(i)}$ can then be written uniformly as
\begin{align}
    A_i
    &=
    \left(r_i-b_{z^{(i)}}\right)
    +
    \lambda\left(b_{z^{(i)}}-\bar{r}\right)
    \nonumber\\
    &=
    r_i
    +
    (\lambda-1)b_{z^{(i)}}
    -
    \lambda\bar{r}.
    \label{eq:unified_advantage}
\end{align}
When $z^{(i)}=\emptyset$, we have $b_{z^{(i)}}=\bar r$, and
\cref{eq:unified_advantage} reduces to $A_i=r_i-\bar r$, consistent with the definition in the main text.

\paragraph{Interpretation as a hierarchical baseline.}
Under Strategy Nudging, rollouts are sampled from a mixture of context-conditioned distributions. The global mean $\bar r$ can therefore be interpreted as an empirical baseline for the overall mixture, whereas $\bar r_z$ serves as a component-specific baseline for rollouts generated under context $z$.

The intra-context term
\begin{equation}
    r_i-b_{z^{(i)}}
\end{equation}
measures the relative quality of a rollout within its assigned context, thereby removing between-context reward variation from the within-context comparison. The context-level term
\begin{equation}
    \lambda\left(b_{z^{(i)}}-\bar r\right)
\end{equation}
then reintroduces a controllable preference based on the relative reward of the corresponding context group.

At $\lambda=1$, the two terms telescope:
\begin{equation}
    A_i
    =
    \left(r_i-b_{z^{(i)}}\right)
    +
    \left(b_{z^{(i)}}-\bar r\right)
    =
    r_i-\bar r.
\end{equation}
Thus, $\lambda=1$ exactly recovers the standard GRPO estimator. Values of $\lambda$ different from one control how strongly context-level reward differences affect credit assignment.

\subsection{Proof of \cref{proposition:3.1}}
\label{apdx:proof_lambda}

Consider two trajectories $y$ and $y'$ with rewards $r$ and $r'$, sampled from context groups $z$ and $z'$, respectively. From \cref{eq:unified_advantage}, their unnormalized advantage difference is
\begin{align}
    A-A'
    &=
    \left[
        r+(\lambda-1)b_z-\lambda\bar r
    \right]
    -
    \left[
        r'+(\lambda-1)b_{z'}-\lambda\bar r
    \right]
    \nonumber\\
    &=
    (r-r')
    +
    (\lambda-1)(b_z-b_{z'}).
    \label{eq:advantage_difference}
\end{align}
The global baseline term cancels because both trajectories belong to the same optimization batch.

Advantage normalization does not affect pairwise ordering. Specifically,
\begin{equation}
    \widehat A-\widehat A'
    =
    \frac{A-A'}{\sigma_A+\delta},
\end{equation}
where $\sigma_A+\delta>0$. Subtracting the common mean $\mu_A$ and dividing by a common positive quantity constitute a positive affine transformation. It therefore suffices to establish the ordering of the unnormalized advantages.

In the binary reward setting, $r,r'\in\{0,1\}$. If $r>r'$, then necessarily $r=1$ and $r'=0$, yielding
\begin{equation}
    A-A'
    =
    1+(\lambda-1)(b_z-b_{z'}).
    \label{eq:binary_advantage_difference}
\end{equation}
Since each effective group baseline is an average of binary rewards,
\begin{equation}
    b_z,b_{z'}\in[0,1],
    \qquad
    |b_z-b_{z'}|\leq1.
\end{equation}

We first consider $\lambda\in(0,1]$. Since $\lambda-1\leq0$, the smallest possible value of \cref{eq:binary_advantage_difference} is attained when $b_z-b_{z'}$ is maximized:
\begin{align}
    A-A'
    &\geq
    1+(\lambda-1)
    \nonumber\\
    &=\lambda
    >0.
    \label{eq:binary_low_lambda}
\end{align}

Next, consider $\lambda\in[1,2)$. Since $\lambda-1\geq0$, the smallest possible value is attained when $b_z-b_{z'}$ is minimized:
\begin{align}
    A-A'
    &\geq
    1-(\lambda-1)
    \nonumber\\
    &=2-\lambda
    >0.
    \label{eq:binary_high_lambda}
\end{align}

It remains to consider the boundary case $\lambda=2$. Because $y$ is a successful trajectory contained in group $z$, its group mean satisfies
\begin{equation}
    b_z>0.
\end{equation}
Similarly, because $y'$ is a failed trajectory contained in group $z'$, its group mean satisfies
\begin{equation}
    b_{z'}<1.
\end{equation}
Consequently, $b_z-b_{z'}>-1$, and
\begin{equation}
    A-A'
    =
    1+(b_z-b_{z'})
    >0.
\end{equation}
Combining the three cases, we obtain
\begin{equation}
    r>r'
    \quad\Longrightarrow\quad
    A>A'
    \qquad
    \text{for all }\lambda\in(0,2].
\end{equation}
Since advantage normalization preserves ordering, the same conclusion holds for $\widehat A$ and $\widehat A'$.\qed

\paragraph{Ordering among equal-reward trajectories.}
When $r=r'$, \cref{eq:advantage_difference} reduces to
\begin{equation}
    A-A'
    =
    (\lambda-1)(b_z-b_{z'}).
    \label{eq:equal_reward_difference}
\end{equation}
Hence, $\lambda$ only changes the relative ordering of trajectories that receive the same reward.

When $\lambda<1$, a trajectory from a lower-reward context receives a larger advantage than an equal-reward trajectory from a higher-reward context. For successful trajectories, this favors solutions discovered under less reliable or less typical contexts, encouraging broader exploration.

When $\lambda>1$, a trajectory from a higher-reward context receives a larger advantage, emphasizing contexts that more consistently induce successful behavior. This enables an intentional separation between exploration and exploitation: Strategy Nudging promotes exploration at the sampling stage by diversifying the context-conditioned rollout distribution, while choosing $\lambda>1$ introduces an exploitation bias in the optimization objective by assigning greater credit to trajectories generated from more reliable contexts. At $\lambda=1$, equal-reward trajectories receive identical advantages regardless of their assigned contexts.

\subsection{Extension to General Reward Functions}
\label{apdx:generalized_reward}

We now extend the analysis to general scalar reward functions. Let
\begin{equation}
    \Delta_{\min}
    =
    \min_{r_i>r_j}(r_i-r_j)
\end{equation}
denote the smallest positive reward gap among the sampled trajectories. We also define the maximum difference between effective group baselines as
\begin{equation}
    D_{\max}
    =
    \max_{z,z'\in\mathcal G(x_0)}
    |b_z-b_{z'}|.
    \label{eq:maximum_baseline_difference}
\end{equation}
We first consider the nontrivial case $D_{\max}>0$. If $D_{\max}=0$, all context groups have the same baseline, and the context-dependent term cannot affect reward ordering for any value of $\lambda$.

For any pair of trajectories satisfying $r>r'$, \cref{eq:advantage_difference} gives
\begin{align}
    A-A'
    &=
    (r-r')
    +
    (\lambda-1)(b_z-b_{z'})
    \nonumber\\
    &\geq
    (r-r')
    -
    |\lambda-1|
    |b_z-b_{z'}|
    \nonumber\\
    &\geq
    \Delta_{\min}
    -
    |\lambda-1|D_{\max}.
    \label{eq:general_reward_lower_bound}
\end{align}
Therefore, a sufficient condition for strict reward-order preservation is
\begin{equation}
    |\lambda-1|
    <
    \frac{\Delta_{\min}}{D_{\max}}.
    \label{eq:general_lambda_condition}
\end{equation}
Under this condition,
\begin{equation}
    r>r'
    \quad\Longrightarrow\quad
    A>A'.
\end{equation}
Because normalization is a positive affine transformation, the ordering is also preserved for the normalized advantages.

The corresponding non-strict condition
\begin{equation}
    |\lambda-1|
    \leq
    \frac{\Delta_{\min}}{D_{\max}}
\end{equation}
guarantees that a higher-reward trajectory never receives a lower advantage, although equality may occur at the boundary.

\section{Details on Baselines}
\label{apdx:baselines}

\paragraph{Rollout Scaling in GRPO.} 
For controlled experiments, we scale the number of rollouts per prompt while adjusting the gradient accumulation steps and generation batch size accordingly, as summarized in \Cref{tab:rl-hparams}. This ensures that the total optimization dynamics remain comparable across different rollout settings.

\paragraph{Example of Generated Contexts.} 
We provide an illustrative example of the strategy-level contexts used in our method. 
These contexts are lightweight, keyword-level hints that do not directly solve the problem, 
but instead steer the model toward distinct reasoning modes. 
Importantly, they are not intended to provide intermediate steps or solutions, 
but rather to act as high-level inductive biases that diversify exploration.

\begin{promptbox}{Problem}
Let $S$ be the set of triples $(a,b,c)$ of non-negative integers such that $a+b+c$ is even. The value of the sum
\[
\sum_{(a,b,c) \in S} \frac{1}{2^a 3^b 5^c}
\]
can be expressed as $\frac{m}{n}$, where $m$ and $n$ are relatively prime positive integers. Compute $m+n$.
\end{promptbox}
\textbf{Strategy-level contexts:}
\begin{promptbox}{}
    "Generating functions", "Parity of sums"
\end{promptbox}

\section{Training Details}
\label{apdx:training}
\paragraph{Framework.}
We used TRL~\citep{trl} to implement the baselines and our algorithm.
\paragraph{Hyperparameters.}
\begin{table}[h]
\caption{Hyperparameters for training.}
\centering
\small
\begin{tabular}{ll}
\toprule
\textbf{Parameter} & \textbf{Value} \\
\midrule
LoRA\citep{hu2022lora} rank & 32 \\
Max prompt length & 2{,}048 \\
Max completion length & 6{,}144\\
RL steps & 500 \\
Batch size & 4 \\
Rollouts per prompt & $N$ \\
Gradient accumulation steps & $\frac{N}{8}$ \\
Generation batch size & $N$\\
Temperature & 1 \\
Min-$p$ & 0.0 \\
Top-$p$ & 0.95 \\
Top-$k$ & $-1$ \\
Learning rate & $2\times10^{-5}$ \\
LR scheduler & cosine \\
Weight decay & 0.001 \\
Warmup ratio & 0.05 \\
Optimizer & AdamW (8-bit) \\
KL coefficient $\beta$ & 0 \\
Epsilon low & 0.2 \\
Epsilon high & 0.2 (unless specified) \\
$\lambda$ & 1.1 \\
$\lambda_\text{PC}$ & 0.1 \\ 
$p_\text{drop}$ & 0.5 \\
Random seed & 42 \\
\bottomrule
\end{tabular}
\label{tab:rl-hparams}
\end{table}
The hyperparameters used in training are given in \Cref{tab:rl-hparams}. $\lambda$ is set to $1.1$ for exploitation among strategies, following the analysis in \Cref{apdx:reward} and \Cref{proposition:3.1}, and $\lambda_\text{PC}$ is set to $0.1$ following \citet{RLTF} without additional hyperparameter optimization. Other non method-specific hyperparameters are shared across \modelname{} and all baselines.

\paragraph{Compute resources.} For all experiments, we used \textsc{NVIDIA H200 140GB GPU}s.

\section{Details on Evaluation}
\label{apdx:eval}

\paragraph{Benchmarks.}
For mathematical reasoning, we evaluate on \texttt{AIME24} and \texttt{AIME25}, 30-problem olympiad-style high-school competitions~\citep{MAA_AIME}; \texttt{AMC23}, a 40-problem high-school contest benchmark~\citep{MAA_AMC}; the level-5 subset of \texttt{MATH500}, containing 134 difficult MATH problems~\citep{MATH500}; and the \texttt{Apex Shortlist}, consisting of 48 advanced competition-style problems~\citep{apex}.
We report \textit{pass@}1 estimated from 128 rollouts using the unbiased estimator of~\citet{passk}, and all solutions are automatically graded using \texttt{math-verify}~\citep{mathverify2024}.

\paragraph{Decoding.}
During evaluation using vLLM~\citep{kwon2025vllm}, all hyperparameters are kept identical to \Cref{tab:rl-hparams}, except for the temperature, which is set to $0.7$.

\section{Details on Efficiency Analysis}
\label{apdx:effie}

This appendix provides the detailed numerical results and computation protocol for the efficiency analysis presented in the main paper.
We analyze both \emph{token efficiency} and \emph{training efficiency} under a unified accounting scheme to enable fair comparison across different exploration methods.

\subsection{Token Efficiency}

\begin{figure}[t]
    \centering
    \includegraphics[width=0.42\textwidth]{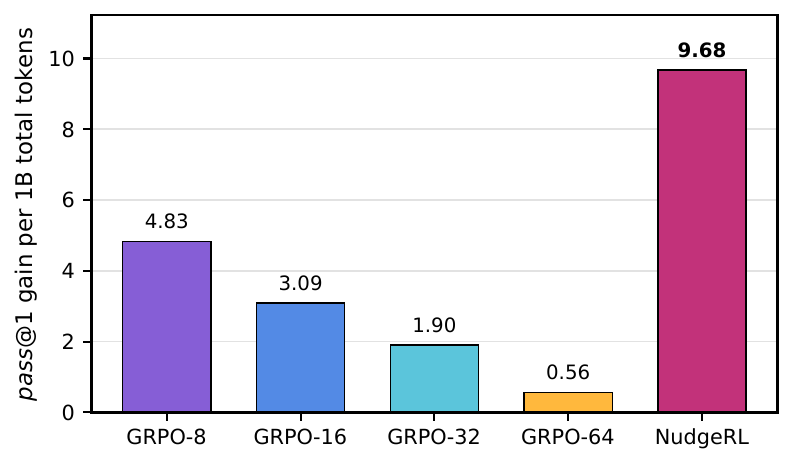}
    \vspace{-0.08in}
    \caption{
    \textbf{Token efficiency.} Average \textit{pass@}1 gain over the base model per 1B total tokens, including the one-time offline context-generation cost of \modelname{}.
    }
    \label{fig:token_efficiency}
    \vspace{-0.18in}
\end{figure}

\Cref{fig:token_efficiency} summarizes token efficiency, and the complete numerical results are reported in \Cref{tab:token_efficiency}.

\begin{table}[h]
\centering
\caption{
\textbf{Token-efficiency comparison on \texttt{Qwen3-4B-Instruct-2507}.}
We report the average \textit{pass@}1 improvement over the base model, the number of training tokens, and the resulting improvement per billion total tokens.
For \modelname{}, total tokens include the one-time offline context-generation cost.
Best results are represented as \textbf{bold}.
}
\label{tab:token_efficiency}
\scriptsize
\setlength{\tabcolsep}{5.5pt}
\renewcommand{\arraystretch}{1.15}
\begin{tabular}{l ccccc}
\toprule
\textbf{Metric}
& \textbf{GRPO-8}
& \textbf{GRPO-16}
& \textbf{GRPO-32}
& \textbf{GRPO-64}
& \textbf{\modelname{}} \\
\midrule

Average \textit{pass@}1 gain
& 0.052
& 0.068
& 0.085
& 0.049
& \textbf{0.087} \\

Completion tokens
& 10{,}455{,}528
& 21{,}497{,}141
& 43{,}590{,}745
& 85{,}450{,}770
& 8{,}523{,}721 \\

Training tokens
& 10{,}762{,}224
& 22{,}034{,}469
& 44{,}665{,}401
& 87{,}600{,}082
& 8{,}860{,}902 \\

Offline context-generation tokens
& 0
& 0
& 0
& 0
& 125{,}000 \\

\cmidrule(lr){1-6}

Total tokens
& 10{,}762{,}224
& 22{,}034{,}469
& 44{,}665{,}401
& 87{,}600{,}082
& 8{,}985{,}902 \\

Average \textit{pass@}1 gain per B total tokens
& 4.83
& 3.09
& 1.90
& 0.56
& \textbf{9.68} \\

\bottomrule
\end{tabular}
\end{table}

For \modelname{}, we separately account for the one-time offline cost of context generation.
Unlike rollout generation during RL training, these contexts are produced only once for each training dataset and can subsequently be reused across different policy initializations, hyperparameter configurations, or repeated training runs.
Consequently, this cost is amortized over all subsequent experiments and does not scale with the number of RL optimization steps.
To provide a conservative comparison, however, all results reported in the main paper include this offline context-generation cost in the total token budget.

\subsection{Training Efficiency}

To compare computational efficiency across methods, we estimate the total training FLOPs using a simplified dense Transformer cost model.
Following standard approximations, we assume a dense 4B-parameter Transformer ($P=4\times10^9$), where

\begin{itemize}
    \item one forward pass over $T$ tokens requires approximately $2PT$ FLOPs,
    \item one forward--backward optimization step requires approximately $6PT$ FLOPs,
    \item autoregressive generation with KV caching requires approximately $2PC$ FLOPs,
\end{itemize}

where $T$ denotes the number of tokens participating in optimization (prompt + completion), and $C$ denotes the generated completion length.

Under this approximation, the per-step computational cost becomes

\begin{align}
\mathrm{GRPO}
&=
2PC
+
6PT,
\\
\mathrm{\modelname{}}
&=
2PC
+
6PT_{x_1}
+
6PT_{x_0},
\end{align}

where $T_{x_0}$ and $T_{x_1}$ denote the optimization tokens for the policy correction objective and the standard RL objective, respectively.

For every training configuration, we record the average prompt length, average completion length, rollout budget, and the total number of optimization steps.
These statistics are then aggregated to estimate the total training compute reported in \Cref{tab:training_flops}.

\begin{table}[h]
\centering
\captionof{table}{
\textbf{Training compute} in EFLOPs ($10^{18}$ FLOPs), split into rollout generation, RL optimization, and policy correction (PC).
}
\label{tab:training_flops}
\vspace{0.03in}

\centering
\scriptsize
\resizebox{0.72\linewidth}{!}{
\setlength{\tabcolsep}{5pt}
\begin{tabular}{l c cccc}
\toprule

\multirow{2}{*}{\textbf{Method}}
& \multirow{2}{*}{\textbf{Rollouts ($N$)}}
& \multicolumn{4}{c}{\textbf{Training Compute (EFLOPs)}} \\

\cmidrule(lr){3-6}

&
& \textbf{Generation}
& \textbf{RL}
& \textbf{PC}
& \textbf{Total} \\

\midrule

\multirow{4}{*}{GRPO}
& 8  & 0.084 & 0.258 & 0 & 0.342 \\
& 16 & 0.172 & 0.529 & 0 & 0.701 \\
& 32 & 0.349 & 1.072 & 0 & 1.421 \\
& 64 & 0.684 & 2.102 & 0 & 2.786 \\

\cmidrule{1-6}

\rowcolor{citeblue!10}
\modelname{}
& 8
& 0.068
& 0.213
& 0.211
& 0.492 \\

\bottomrule
\end{tabular}
}
\end{table}

Note that our estimates intentionally exclude infrastructure-dependent overheads such as optimizer state updates, communication costs, activation checkpointing, and data loading, as these are shared across methods and do not materially affect the relative comparison.
Accordingly, the reported EFLOPs should be interpreted as normalized estimates of model computation rather than exact hardware-level measurements.

\section{Results across Model Families}
\label{apdx:model}

\Cref{tab:main_table_v3} reports the full main results on both model families.
On \texttt{Olmo3-7B-Instruct-SFT}~\citep{Olmo}, \modelname{} with 8 rollouts achieves the best average \textit{pass@}1 of 0.285, slightly above the strongest GRPO baseline (0.281 at $N=32$) and clearly above GRPO at the matched budget of $N=8$ (0.268).
The trend therefore mirrors the one observed on \texttt{Qwen3-4B-Instruct-2507}: improving the \emph{quality} of exploration is more effective than scaling the rollout budget, and the benefit is not specific to a single pretraining recipe.

We also observe that GRPO at $N=64$ collapses on \texttt{Olmo3-7B-Instruct-SFT} (0.160, well below the base model at 0.225), a more severe version of the degradation seen on \texttt{Qwen3-4B-Instruct-2507}.
This is consistent with the instability of large-batch rollout scaling discussed by~\citet{BRORL}.

\begin{table}[t]
\centering
\caption{
\textbf{Full results across both model families}, comparing rollout scaling and context-based exploration. We report \textit{pass@}1 estimated using 128 rollouts. Best results are \textbf{bold} and second best \underline{underlined}. The \texttt{Qwen3-4B-Instruct} block reproduces \Cref{tab:main_table}.
}

\label{tab:main_table_v3}
\small
\resizebox{\textwidth}{!}{
\setlength{\tabcolsep}{5.5pt}
\begin{tabular}{c l c ccccccc}
\toprule
\textbf{Model} & \textbf{Method} & \textbf{Rollouts ($N$)} 
& \textbf{AIME24} & \textbf{AIME25} 
& \textbf{AMC23} & \textbf{MATH500} & \textbf{APEX} & \textbf{Average} \\

\midrule

\multirow{7}{*}[-2.5ex]{\texttt{\shortstack{Qwen3-4B-\\Instruct}}}
& Base model 
& -- 
& 0.374 & 0.352 & 0.653 & 0.592 & 0.036 & 0.402 \\

\cmidrule{2-9}
& \multirow{4}{*}{GRPO}
& 8 
& 0.444 & 0.367 & 0.749 & \underline{0.668} & 0.040 & 0.454 \\
& 
& 16 
& 0.454 & 0.355 & 0.840 & 0.655 & 0.045 & 0.470 \\
& 
& 32 
& 0.451 & \underline{0.370} & \textbf{0.881} & \textbf{0.674} & \textbf{0.058} & \underline{0.487} \\
& 
& 64 
& 0.415 & 0.324 & 0.848 & 0.641 & 0.027 & 0.451 \\

\cmidrule{2-9}
\rowcolor{citeblue!10}
\cellcolor{white}
& \modelname{}
& 8 
& \textbf{0.482} & \textbf{0.393} & \underline{0.857} & 0.660 & \underline{0.053} & \textbf{0.489} \\

\midrule

\multirow{7}{*}[-2.5ex]{\texttt{\shortstack{Olmo3-7B-\\Instruct-SFT}}}
& Base model 
& -- 
& 0.134 & 0.118 & 0.467 & 0.384 & 0.021 & 0.225 \\

\cmidrule{2-9}
& \multirow{4}{*}{GRPO}
& 8 
& 0.187 & 0.159 & 0.537 & 0.434 & \underline{0.025} & 0.268 \\
& 
& 16 
& 0.188 & 0.176 & 0.548 & \underline{0.461} & 0.023 & 0.279 \\
& 
& 32 
& \textbf{0.195} & \underline{0.176} & 0.553 & 0.459 & 0.024 & \underline{0.281} \\
& 
& 64 
& 0.081 & 0.053 & 0.349 & 0.291 & \textbf{0.027} & 0.160 \\

\cmidrule{2-9}
\rowcolor{citeblue!10}
\cellcolor{white}
& \modelname{}
& 8 
& \underline{0.190} & \textbf{0.179} & \textbf{0.563} & \textbf{0.468} & \underline{0.025} & \textbf{0.285} \\

\bottomrule
\end{tabular}
}
\vspace{-0.2in}
\end{table}

\section{Results on Code Generation}
\label{apdx:code}

To demonstrate that \modelname{} generalizes beyond mathematical reasoning, we train \texttt{Qwen2.5-Coder-7B-Instruct}~\citep{QwenCoder} on \texttt{LiveCodeBench v1--v5}~\citep{LiveCodeBench} and evaluate on \texttt{v6}. As shown in \Cref{tab:livecodebench_v6}, \modelname{} consistently outperforms GRPO across all rollout budgets while using only 8 rollouts, achieving the best overall \textit{pass@}1 of 0.335 compared to the strongest GRPO baseline of 0.317 at 64 rollouts. The improvement is consistent across both AtCoder and LeetCode subsets, suggesting that the benefit of Strategy Nudging extends beyond mathematical reasoning to code generation, where improving exploration quality is again more effective than simply scaling rollout budgets.

\begin{table}[t]
\centering
\caption{
\textbf{Results on LiveCodeBench v6} comparing rollout scaling
and context-based exploration.
We report \textit{pass@}1 estimated using 128 rollouts,
including both the per-source breakdown and the overall performance.
Best results are shown in \textbf{bold}, and second-best results are
\underline{underlined}.
}
\vspace{0.02in}
\label{tab:livecodebench_v6}
\small
\resizebox{\textwidth}{!}{
\setlength{\tabcolsep}{7pt}
\begin{tabular}{c l c cc c}
\toprule

\textbf{Model}
& \textbf{Method}
& \textbf{Rollouts ($N$)}
& \multicolumn{2}{c}{\textbf{Per-Source Breakdown}}
& \multicolumn{1}{c}{\textbf{LiveCodeBench v6}} \\

\cmidrule(lr){4-5}
\cmidrule(lr){6-6}

&
&
& \textbf{AtCoder \textit{pass@}1}
& \textbf{LeetCode \textit{pass@}1}
& \textbf{Total \textit{pass@}1} \\

\midrule

\multirow{6}{*}[-2.0ex]{
\texttt{\shortstack{Qwen2.5-Coder-\\7B-Instruct}}
}
& Base model
& --
& 0.325
& 0.254
& 0.299 \\

\cmidrule{2-6}

& \multirow{4}{*}{GRPO}
& 8
& 0.325
& 0.276
& 0.308 \\

&
& 16
& 0.332
& 0.285
& 0.315 \\

&
& 32
& \underline{0.335}
& 0.277
& 0.314 \\

&
& 64
& 0.334
& \underline{0.287}
& \underline{0.317} \\

\cmidrule{2-6}

\rowcolor{citeblue!10}
\cellcolor{white}
& \modelname{}
& 8
& \textbf{0.352}
& \textbf{0.304}
& \textbf{0.335} \\

\bottomrule
\end{tabular}
}
\vspace{0.2in}
\end{table}

\section{Training and Inference-Time Dynamics}
\label{apdx:dynamics}

\begin{figure}[t]
    \centering
    \begin{subfigure}[b]{0.31\textwidth}
        \includegraphics[width=\linewidth]{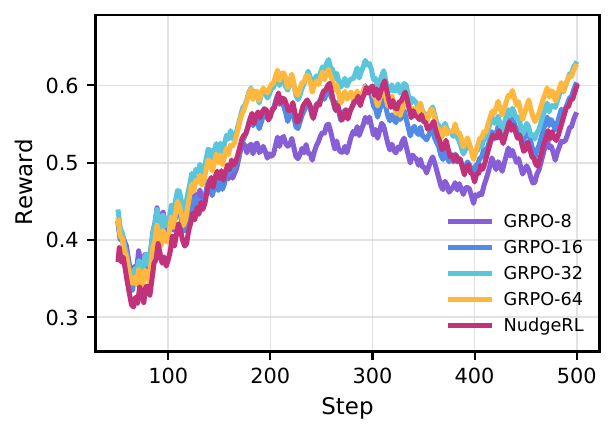}
        \caption{Training reward}\label{fig:reward}
    \end{subfigure}\hfill
    \begin{subfigure}[b]{0.31\textwidth}
        \includegraphics[width=\linewidth]{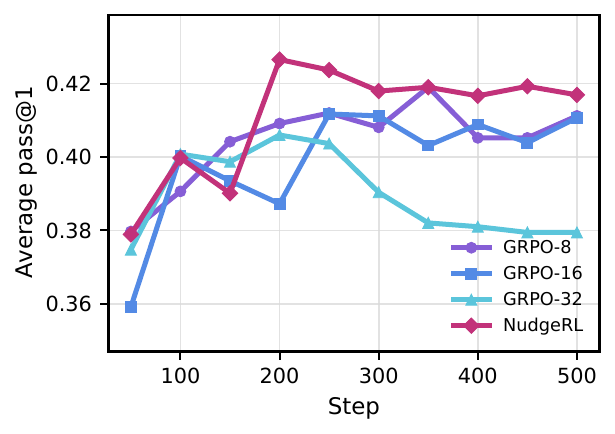}
        \caption{\textit{pass@}1}\label{fig:pass1}
    \end{subfigure}\hfill
    \begin{subfigure}[b]{0.31\textwidth}
        \includegraphics[width=\linewidth]{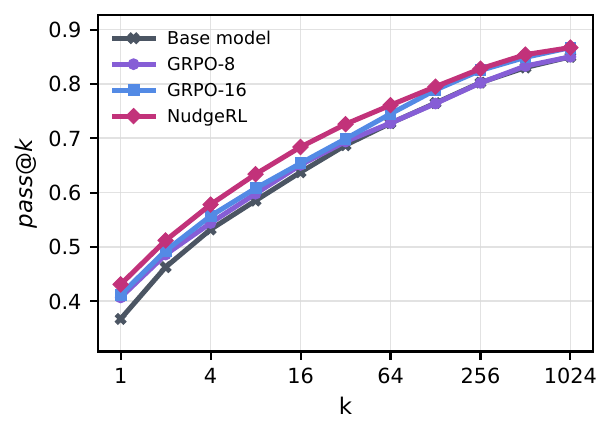}
        \caption{\textit{pass@}k}\label{fig:passk}
    \end{subfigure}
    \vspace{-0.05in}
    \caption{\textbf{Training and inference-time behavior} on \texttt{AIME24/25}.
    (a) EMA-smoothed training reward (decay 0.99). (b) \emph{pass@1} during training.
    (c) Average \emph{pass@k}.}
    \label{fig:training_dynamics}
    \vspace{-0.15in}
\end{figure} 

As shown in~\Cref{fig:pass1}, \modelname{} improves \textit{pass@}1 substantially faster than the GRPO variants and remains the strongest method throughout most of training. This suggests that Strategy Nudging improves sample efficiency by exposing useful reasoning trajectories earlier, rather than merely increasing the number of sampled rollouts.

This trend also extends to inference-time sampling. As shown in~\Cref{fig:passk}, \modelname{} consistently outperforms GRPO-8 across the full range of sampling budgets, indicating that it requires fewer generated solutions to achieve the same level of \textit{pass@}k.

\begin{figure}
    \centering
    \includegraphics[width=1.0\linewidth]{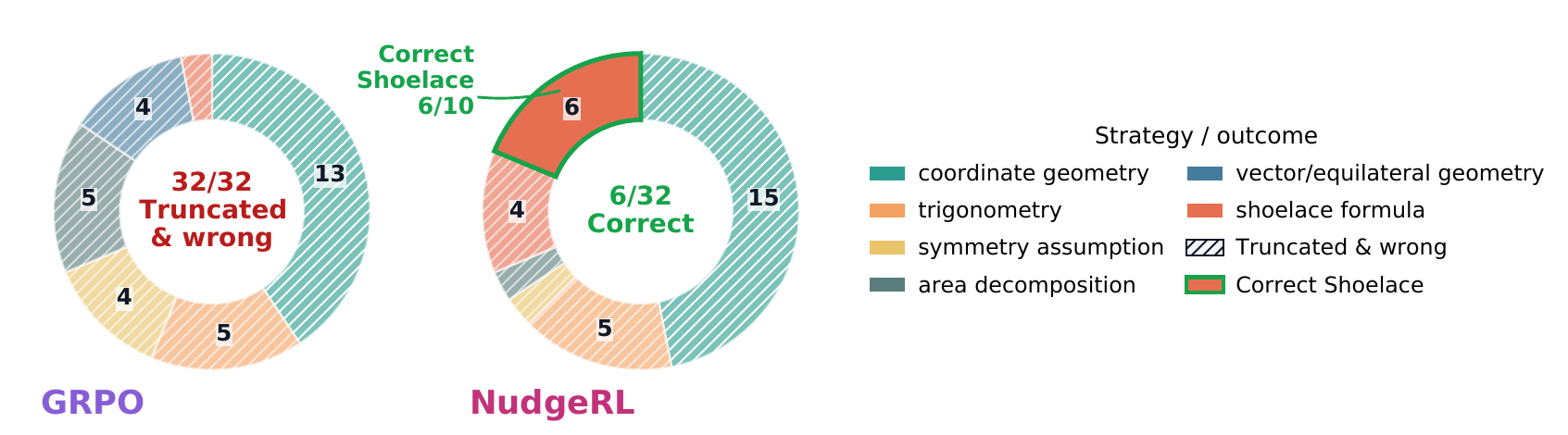}
    \vspace{-0.2in}
    \caption{\textbf{\modelname{} internalizes effective test-time strategies.} Across 32 rollouts on a \texttt{AIME25} problem, GRPO yields only incorrect and truncated trajectories. Conversely, \textsc{NudgeRL} produces 6 correct solutions using the shoelace formula.}
    % demonstrating that training strategies successfully transfer to the base policy without needing external test-time contexts.}
    \label{fig:casestudy}
    %\vspace{-0.1in}
\end{figure}
\section{Case Study}
\label{apdx:case}

To examine the source of the performance gains of \modelname{}, we analyze an \texttt{AIME25} problem (shown below) on which the \modelname{}-trained model sampled correct trajectories while the GRPO-trained model failed on every rollout. For each model, we sampled 32 rollouts and categorized each by its dominant reasoning strategy.

\begin{promptbox}{Problem}
    Let $\triangle ABC$ be a right triangle with $\angle A = 90^\circ$ and $BC = 38$. There exist points $K$ and $L$ inside the triangle such that
    \[
    AK = AL = BK = CL = KL = 14.
    \]
    The area of the quadrilateral $BKLC$ can be expressed as $n\sqrt{3}$ for some positive integer $n$. Find $n$.
\end{promptbox}

As shown in \Cref{fig:casestudy}, both models relied predominantly on \emph{coordinate geometry}. The GRPO-trained model, however, additionally explored heuristic \emph{symmetry assumptions} and case-by-case \emph{area decomposition}. Although these strategies occasionally progressed toward partial solutions, they led to long derivations that were truncated before reaching a final answer, and all 32 trajectories failed. The \emph{shoelace formula}, which computes the area of a polygon directly from its vertex coordinates, appeared in only one GRPO rollout. In contrast, \modelname{} sampled this strategy far more often, and it yielded substantially shorter derivations that reached the correct answer within the generation budget.

This behavior is consistent with the complementary roles of the two components of our framework: \emph{Strategy Nudging} exposes rare but effective reasoning modes such as the shoelace formula, while the \emph{Inter-Intra Group Advantage} reinforces and exploits such strategies once they are discovered.

\section{Mechanism Analyses and Full Evaluation Results}
\label{apdx:ablation}

To understand the source of performance gains in \modelname{}, we conduct a series of controlled experiments using \texttt{Qwen3-4B-Instruct-2507} on a subset of benchmarks.

\begin{figure}[t]
    \centering
    \begin{subfigure}[b]{0.31\textwidth}
        \includegraphics[width=\linewidth]{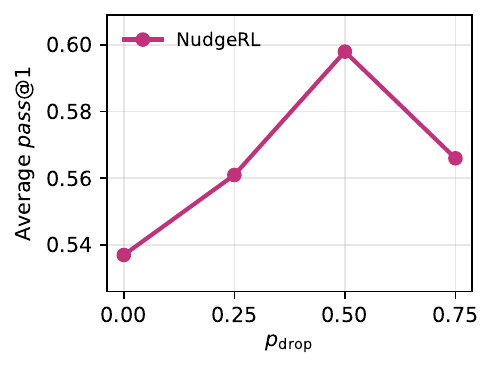}
        \caption{Ablation on $p_\text{drop}$}\label{fig:p_drop}
    \end{subfigure}\hfill
    \begin{subfigure}[b]{0.31\textwidth}
        \includegraphics[width=\linewidth]{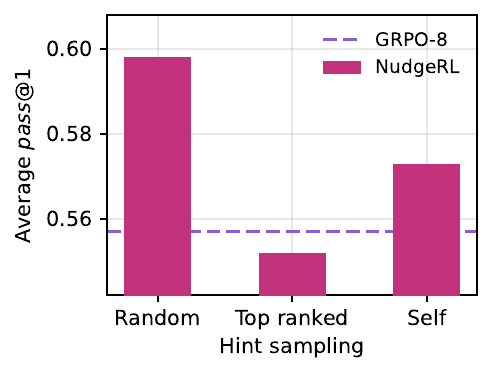}
        \caption{Ablation on sampling}\label{fig:sampling}
    \end{subfigure}\hfill
    \begin{subfigure}[b]{0.31\textwidth}
        \includegraphics[width=\linewidth]{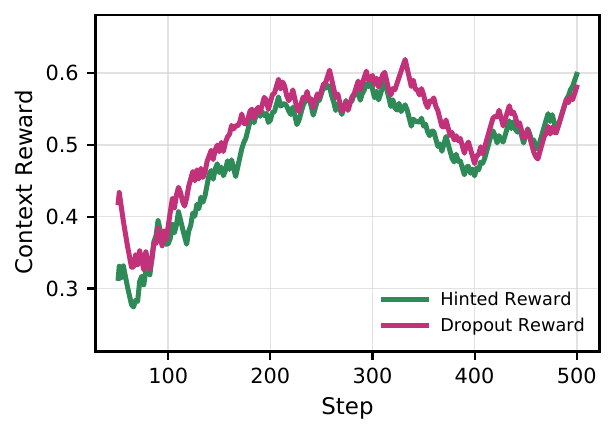}
        \caption{Context reward}\label{fig:context_reward}
    \end{subfigure}
    \vspace{-0.05in}
    \caption{\textbf{Mechanism analyses.} (a, b) Average \textit{pass@}1 on \texttt{AIME24/25, AMC23, MATH500}. (c) Time-weighted EMA reward mean (0.99)
    with and without context.}
    \label{fig:mechanism}
    \vspace{-0.1in}
\end{figure}

\paragraph{$p_\textrm{drop}$ ablation.}
As shown in \Cref{fig:p_drop} and \Cref{tab:pdrop_table_v2}, a moderate dropout rate ($p_\textrm{drop}=0.5$) consistently yields the best performance across benchmarks.
Context dropout plays a dual role: it enables exploration beyond fixed contexts by occasionally reverting to the base prompt, while also stabilizing group-wise statistics through a more balanced sample distribution.
When $p_\textrm{drop}=0$, exploration is restricted to predefined contexts, whereas large values diminish the influence of context forcing.
These results suggest that maintaining a balanced mixture of context-conditioned and context-free samples is important for achieving both diverse exploration and stable optimization.

\paragraph{Hint sampling.}
We also study how the quality of sampled contexts affects performance by comparing three strategies: \textit{random} sampling, \textit{top-ranked} selection, and \textit{self}-generated contexts.
In the top-ranked setting, we first generate a pool of five candidate contexts and select the two that yield the largest improvement in \textit{pass@}16 for each problem. For self-generated contexts, we prompt the base model to produce its own strategy hints.

As shown in \Cref{fig:sampling} and \Cref{tab:hint_table_v2}, random sampling consistently achieves the best \textit{pass@}1, outperforming both top-ranked selection and self-generated contexts. While top-ranked contexts ensure higher immediate correctness, utilizing them to \textit{nudge} the training policy yields lower performance than the GRPO baseline. Self-generated contexts improve over the GRPO baseline, but remain less effective than randomly sampled external contexts, likely because they inherit the policy's existing exploration bias.
\paragraph{Effect of contexts during training.}
We also report the dropout reward mean ($\mathbb{E}_i[r(x_0,y^{(i)}) \mid z^{(i)}=\emptyset]$) and the hinted reward mean ($\mathbb{E}_i[r(x_0,y^{(i)}) \mid z^{(i)}\neq\emptyset]$) during training with \modelname{}. As shown in \Cref{fig:context_reward}, both rewards improve together throughout training, suggesting that trajectories discovered under context-conditioned exploration are successfully transferred to the base policy through the policy correction objective.
Interestingly, the dropout reward occasionally exceeds the hinted reward during training. This contrasts with prior feasibility-oriented methods based on privileged information~\citep{POPE, chen2025nudging}.

These results suggest that, within our framework, the primary role of context is not to provide the single best hint, but to promote diversity in reasoning.

\subsection{Full Evaluation Results}
\begin{table}[h]
\centering
\caption{
\textbf{$p_\textrm{drop}$ ablation results.} We report \textit{pass@}1 estimated using 128 rollouts. Best results are represented as \textbf{bold}.
}
\label{tab:pdrop_table_v2}
\scriptsize
\setlength{\tabcolsep}{6pt}
\renewcommand{\arraystretch}{1.15}
\begin{tabular}{lccccc}
\toprule
$p_\textrm{drop}$
& \textbf{AIME24}
& \textbf{AIME25}
& \textbf{AMC23}
& \textbf{MATH500}
& \textbf{Average} \\
\midrule

0.00
& 0.418
& 0.344
& 0.759
& 0.628
& 0.537 \\

0.25
& 0.461
& 0.354
& 0.773
& 0.658
& 0.561 \\

0.50
& \textbf{0.482}
& \textbf{0.393}
& \textbf{0.857}
& \textbf{0.660}
& \textbf{0.598} \\

0.75
& 0.458
& 0.361
& 0.796
& 0.649
& 0.566 \\

\bottomrule
\end{tabular}
\end{table}
\begin{table}[h]
\centering
\caption{
\textbf{Hint sampling ablation results.} We report \textit{pass@}1 estimated using 128 rollouts. Best results are represented as \textbf{bold}.
}
\label{tab:hint_table_v2}
\scriptsize
\setlength{\tabcolsep}{6pt}
\renewcommand{\arraystretch}{1.15}
\begin{tabular}{lccccc}
\toprule
Sampling
& \textbf{AIME24}
& \textbf{AIME25}
& \textbf{AMC23}
& \textbf{MATH500}
& \textbf{Average} \\
\midrule
GRPO(8 rollouts, baseline)
& 0.444
& 0.367
& 0.749
& 0.668
& 0.557 \\
Random
& \textbf{0.482}
& \textbf{0.393}
& \textbf{0.857}
& \textbf{0.660}
& \textbf{0.598} \\

Top ranked
& 0.448
& 0.355
& 0.774
& 0.632
& 0.552 \\

Self
& 0.436
& 0.373
& 0.844
& 0.639
& 0.573 \\

\bottomrule
\end{tabular}
\end{table}

\section{Additional Experiments}
\label{apdx:additional}
\subsection{Comparison with Oracle Prefix Baseline}

To compare \modelname{} with methods that leverage privileged information, we implement an oracle prefix baseline following POPE~\citep{POPE}. Specifically, we adopt the prompt format and dataset mixture (i.e., with and without oracle solutions) from the original work. Since the adaptive prefix-length selection strategy used in POPE is not publicly available, we instead evaluate a range of fixed prefix ratios. Oracle solutions are generated using \texttt{DeepSeek-R1-0528}~\citep{DS}, and an example oracle solution for the problem presented in \cref{apdx:baselines} is shown below.

\textbf{Oracle solution:}
\begin{promptbox}{}
    We compute the sum
    \[
    \sum_{(a,b,c) \in S} \frac{1}{2^a 3^b 5^c},
    \]
    where \(S\) consists of non-negative integer triples \((a,b,c)\) with \(a+b+c\) even. Using the indicator
    \[
    \mathbf{1}_{a+b+c \text{ even}} = \frac{1}{2}\left(1 + (-1)^{a+b+c}\right),
    \]
    we write
    \[
    \sum_{(a,b,c) \in S} \frac{1}{2^a 3^b 5^c}
    =
    \frac{1}{2}
    \left(
    \sum_{a,b,c \geq 0} \frac{1}{2^a 3^b 5^c}
    +
    \sum_{a,b,c \geq 0} \frac{(-1)^{a+b+c}}{2^a 3^b 5^c}
    \right).
    \]

    The first sum factors as
    \[
    \sum_{a \geq 0} \frac{1}{2^a}
    \cdot
    \sum_{b \geq 0} \frac{1}{3^b}
    \cdot
    \sum_{c \geq 0} \frac{1}{5^c}
    =
    \frac{1}{1 - \frac{1}{2}}
    \cdot
    \frac{1}{1 - \frac{1}{3}}
    \cdot
    \frac{1}{1 - \frac{1}{5}}
    =
    2 \cdot \frac{3}{2} \cdot \frac{5}{4}
    =
    \frac{15}{4}.
    \]

    The second sum factors as
    \[
    \sum_{a \geq 0} \frac{(-1)^a}{2^a}
    \cdot
    \sum_{b \geq 0} \frac{(-1)^b}{3^b}
    \cdot
    \sum_{c \geq 0} \frac{(-1)^c}{5^c}
    =
    \frac{1}{1 + \frac{1}{2}}
    \cdot
    \frac{1}{1 + \frac{1}{3}}
    \cdot
    \frac{1}{1 + \frac{1}{5}}
    =
    \frac{2}{3} \cdot \frac{3}{4} \cdot \frac{5}{6}
    =
    \frac{5}{12}.
    \]

    Thus, the total sum is
    \[
    \frac{1}{2}
    \left(
    \frac{15}{4} + \frac{5}{12}
    \right)
    =
    \frac{1}{2} \cdot \frac{50}{12}
    =
    \frac{25}{12}.
    \]

    Since \(\frac{25}{12}\) is in lowest terms, we have \(m = 25\), \(n = 12\), and \(m + n = 37\).
\end{promptbox}

We train \texttt{Qwen3-4B-Instruct-2507} using prefix ratios
$\{0.05, 0.10, 0.15, 0.20, 0.25\}$.
The results are summarized in \cref{tab:oracle_prefix}.
\begin{table}[h]
\centering
\caption{
\textbf{Comparison with oracle prefix baseline.} We report \textit{pass@}1 estimated using 128 rollouts. Best results are represented as \textbf{bold}.
}
\label{tab:oracle_prefix}
\scriptsize
\setlength{\tabcolsep}{6pt}
\renewcommand{\arraystretch}{1.15}
\begin{tabular}{c cccc}
\toprule
\textbf{Method}
& \textbf{Prefix Ratio}
& \textbf{AIME24}
& \textbf{AIME25}
& \textbf{Average} \\
\midrule

\multirow{5}{*}{Oracle Prefix}
& 0.05
& 0.404
& 0.342
& 0.373 \\

& 0.10
& 0.405
& \textbf{0.369}
& 0.387 \\

& 0.15
& \textbf{0.460}
& 0.337
& 0.399 \\

& 0.20
& 0.439
& 0.364
& \textbf{0.401} \\

& 0.25
& 0.430
& 0.368
& 0.399 \\

\cmidrule(lr){1-5}

Best Oracle Prefix
& --
& \textbf{0.460}
& \textbf{0.369}
& \textbf{0.415} \\

\midrule

NudgeRL
& --
& \textbf{0.482}
& \textbf{0.393}
& \textbf{0.438} \\

\bottomrule
\end{tabular}
\end{table}

To provide an optimistic upper bound, we additionally report \textit{Best Oracle Prefix}, which selects the best-performing prefix ratio independently for each benchmark. Even under this oracle selection protocol, the average performance reaches only 0.415, remaining below \modelname{} (0.438). Furthermore, \modelname{} consistently outperforms the best oracle-prefix configuration on both AIME24 (0.482 vs.\ 0.460) and AIME25 (0.393 vs.\ 0.369).

These results indicate that the advantage of \modelname{} cannot be explained by a particular choice of oracle prefix length. Rather than exposing a partial ground-truth solution, encouraging exploration through diverse reasoning contexts proves more effective for improving downstream policy optimization.

Interestingly, \modelname{} also does not require the context generator itself to solve the target problem. In our experiments, the context generator (\texttt{gpt-4o-mini}) achieves \textit{pass@}16 = 0 on approximately 33\% of the 200 \texttt{DAPO-17k-Processed} problems. Nevertheless, the resulting contexts still substantially improve downstream exploration and ultimately outperform oracle-prefix training. This observation suggests that the role of contextual hints is fundamentally different from that of oracle solutions: they need not encode the correct reasoning process, but only provide sufficiently diverse and plausible strategic directions that guide the policy toward exploring reasoning trajectories it would otherwise fail to discover.
\subsection{Comparison with $\epsilon_\textrm{high}$ scaling.}

We further compare our algorithm with decoupled clipping~\citep{DAPO}:
$\mathrm{clip}(r, 1-\epsilon_\textrm{low}, 1+\epsilon_\textrm{high}),$
where $\epsilon_\textrm{high}$ controls the strength of policy updates by amplifying the contribution of successful trajectories. Increasing $\epsilon_\textrm{high}$ therefore allows more aggressive policy updates toward positive-advantage trajectories.
As shown in \Cref{tab:eps_table_v2}, increasing $\epsilon_\textrm{high}$ generally improves GRPO performance in the moderate regime used in prior works~\citep{GRPO, DAPO}. However, our method with $\epsilon_\textrm{high}=0.2$ consistently outperforms GRPO across the entire scaling range from moderate to extreme values. This suggests that improving exploration quality is more effective than simply increasing the magnitude of stochastic policy updates. Additionally, under the more extreme scaling adopted in recent RLVR settings~\citep{prorl}, GRPO sharply deteriorates at $\epsilon_\textrm{high}=0.4$. We argue that this degradation highlights a limitation of purely stochastic distribution-level exploration: increasing update magnitude alone provides little control over \emph{what} is explored. 
\begin{table}[h]
\centering
\caption{
$\epsilon_\textrm{high}$ scaling results. We report \textit{pass@}1 estimated using 128 rollouts. Best results are represented as \textbf{bold}.
}
\label{tab:eps_table_v2}
\scriptsize
\setlength{\tabcolsep}{6pt}
\renewcommand{\arraystretch}{1.15}
\begin{tabular}{c ccccccc}
\toprule
\textbf{Algorithm} 
& $\epsilon_\textrm{high}$
& \textbf{AIME24}
& \textbf{AIME25}
& \textbf{AMC23}
& \textbf{MATH500}
& \textbf{Average} \\
\midrule

NudgeRL
& 0.2
& \textbf{0.482}
& \textbf{0.393}
& \textbf{0.857}
& \textbf{0.660}
& \textbf{0.598} \\

\midrule

\multirow{6}{*}{GRPO}
& 0.2
& 0.444
& 0.367
& 0.749
& \textbf{0.668}
& 0.557 \\

& 0.24
& 0.451
& \textbf{0.373}
& 0.795
& 0.645
& \textbf{0.566} \\

& 0.28
& 0.443
& 0.372
& 0.793
& 0.648
& 0.564 \\

& 0.32
& \textbf{0.452}
& 0.358
& 0.813
& 0.640
& \textbf{0.566} \\

& 0.36
& 0.432
& 0.338
& \textbf{0.845}
& 0.647
& 0.565 \\

& 0.40
& 0.406
& 0.341
& 0.781
& 0.638
& 0.541 \\

\bottomrule
\end{tabular}
\end{table}
\subsection{Ablations on the Learning Objective}
We ablate the two coefficients of the \modelname{} objective, again using \texttt{Qwen3-4B-Instruct-2507} on a subset of benchmarks.

\paragraph{Exploration-exploitation trade-off via $\lambda$.}

\Cref{tab:lbd_table_v2} presents the effect of varying $\lambda$, where $\lambda=1.1$  achieves the best performance.
This trend aligns with our Proposition~\ref{proposition:3.1} in~\Cref{sec:interintra}. Since strategy nudging already ensures sufficient diversity at the sampling stage, increasing $\lambda$ does not hinder exploration across contexts. Instead, it strengthens exploitation within each problem by prioritizing trajectories from more reliable contexts. This leads to more consistent learning of high-quality solutions per instance, explaining the observed performance gains at $\lambda=1.1$.
\begin{table}[h]
\centering
\caption{
\textbf{$\lambda$ ablation results.} We report \textit{pass@}1 estimated using 128 rollouts. Best results are represented as \textbf{bold}.
}
\label{tab:lbd_table_v2}
\scriptsize
\setlength{\tabcolsep}{6pt}
\renewcommand{\arraystretch}{1.15}
\begin{tabular}{lccccc}
\toprule
$\lambda$
& \textbf{AIME24}
& \textbf{AIME25}
& \textbf{AMC23}
& \textbf{MATH500}
& \textbf{Average} \\
\midrule

0.9
& 0.403
& 0.366
& 0.806
& 0.648
& 0.556 \\

1.0
& 0.436
& 0.359
& 0.831
& 0.643
& 0.567 \\

1.1
& \textbf{0.482}
& \textbf{0.393}
& \textbf{0.857}
& \textbf{0.660}
& \textbf{0.598} \\

\bottomrule
\end{tabular}
\end{table}

\paragraph{Policy Correction Coefficient.}
As shown in \Cref{tab:distill_table_v2}, removing the policy correction term ($\lambda_\textrm{PC}=0$) results in a clear performance drop, indicating that explicitly transferring context-discovered trajectories to the base policy is essential. 
However, overly large values also degrade performance, likely due to over-constraining the policy toward sampled trajectories. 
A moderate coefficient ($\lambda_\textrm{PC}=0.1$) achieves the best results, suggesting that policy correction should complement the underlying RL objective.
\begin{table}[h]
\centering
\caption{
\textbf{$\lambda_\textrm{PC}$ ablation results.} We report \textit{pass@}1 estimated using 128 rollouts. Best results are represented as \textbf{bold}.
}
\label{tab:distill_table_v2}
\scriptsize
\setlength{\tabcolsep}{6pt}
\renewcommand{\arraystretch}{1.15}
\begin{tabular}{lccccc}
\toprule
$\lambda_\textrm{PC}$
& \textbf{AIME24}
& \textbf{AIME25}
& \textbf{AMC23}
& \textbf{MATH500}
& \textbf{Average} \\
\midrule

0.0
& 0.423
& 0.362
& 0.826
& 0.628
& 0.560 \\

0.1
& \textbf{0.482}
& \textbf{0.393}
& \textbf{0.857}
& \textbf{0.660}
& \textbf{0.598} \\

0.5
& 0.425
& 0.361
& 0.730
& 0.629
& 0.536 \\

\bottomrule
\end{tabular}
\end{table}

\section{Broader Impacts}
\label{apdx:bi}

This paper proposes an efficient framework for structured exploration in reinforcement learning with verifiable rewards (RLVR). On the positive side, our method improves exploration efficiency without relying on extremely large rollout budgets or expensive oracle supervision, which may help reduce the computational cost of training reasoning models and improve accessibility for smaller research groups.

However, improving exploration efficiency may also contribute to the development of increasingly capable reasoning systems, which could be misused in harmful or unintended ways. We therefore emphasize the importance of continued research on safety, oversight, and responsible deployment.

\end{document}